\documentclass[conference]{IEEEtran}
\IEEEoverridecommandlockouts
\usepackage{amsmath,amssymb,amsfonts}
\usepackage{algorithmic}
\usepackage[linesnumbered,ruled]{algorithm2e}
\usepackage{graphicx}
\usepackage{textcomp}
\usepackage{xcolor}
\usepackage[colorlinks=true, urlcolor=blue]{hyperref}
\usepackage{enumitem}
\usepackage{tabularx}
\usepackage{booktabs}   
\usepackage[
    backend=biber,
    style=ieee,
    sorting=none,
    maxnames=6,
    minnames=6
]{biblatex}

\def\BibTeX{{\rm B\kern-.05em{\sc i\kern-.025em b}\kern-.08em
    T\kern-.1667em\lower.7ex\hbox{E}\kern-.125emX}}

\begin{document}

\title{
ColLab: A Collaborative Spatial Progressive \\ 
Data Engine for Referring Expression \\
Comprehension and Generation
}
\makeatletter
\newcommand{\linebreakand}{
  \end{@IEEEauthorhalign}
  \hfill\mbox{}\par
  \mbox{}\hfill\begin{@IEEEauthorhalign}
}
\makeatother

\author{
Shilan Zhang\textsuperscript{1}
\and
Jirui Huang\textsuperscript{1}
\and
Ruilin Yao\textsuperscript{1}
\and 
Cong Wang\textsuperscript{2}
\and
Yaxiong Chen\textsuperscript{1}
\and
Peng Xu\textsuperscript{3}
\and
Shengwu Xiong\textsuperscript{1} 
\and
\centerline{\textsuperscript{1} Wuhan University of Technology; 
\textsuperscript{2} Northwestern Polytechnical University; \textsuperscript{3} Tsinghua University}\\
~\\
Project page: \href{https://github.com/mars2workshop/ColLab}{https://github.com/mars2workshop/ColLab} \\
ICCV 2025 MARS2 workshop \& challenge website:
\href{https://mars2workshop.github.io/iccv2025/}{https://mars2workshop.github.io/iccv2025/}
}

\maketitle

\begin{abstract}
Referring Expression Comprehension (REC) and Referring Expression Generation (REG) are fundamental tasks in multimodal understanding, supporting precise object localization through natural language.
However, existing REC and REG datasets rely heavily on manual annotation, which is labor-intensive and difficult to scale.
In this paper, we propose ColLab, a collaborative spatial progressive data engine that enables fully automated REC and REG data generation without human supervision.
Specifically, our method introduces a Collaborative Multimodal Model Interaction (CMMI) strategy, which leverages the semantic understanding of multimodal large language models (MLLMs) and large language models (LLMs) to generate descriptions.
Furthermore, we design a module termed Spatial Progressive Augmentation (SPA) to enhance spatial expressiveness among duplicate instances. 
Experiments demonstrate that ColLab significantly accelerates the annotation process of REC and REG while improving the quality and discriminability of the generated expressions. In addition to the core methodological contribution, our framework was partially adopted in the data generation pipeline of the ICCV 2025 MARS2 Challenge on Multimodal Reasoning, enriching the dataset with diverse and challenging samples that better reflect real-world reasoning demands.

\end{abstract}


\section{Introduction}
Referring Expression Comprehension (REC)~\cite{rec_yao2024visual,rec_reg_ma2024groma} and Referring Expression Generation (REG)~\cite{reg_ye2023whether} are fundamental tasks that connect visual content with natural language. These tasks play a crucial role in a wide range of applications such as human-computer interaction, autonomous navigation, and visual question answering. 
However, manual annotation of REC and REG data is notoriously labor-intensive, time-consuming, and prone to inconsistencies, which limits the scalability of existing datasets and impedes further progress.

\begin{figure}[!h]
\centerline{\includegraphics[width=0.5\textwidth]{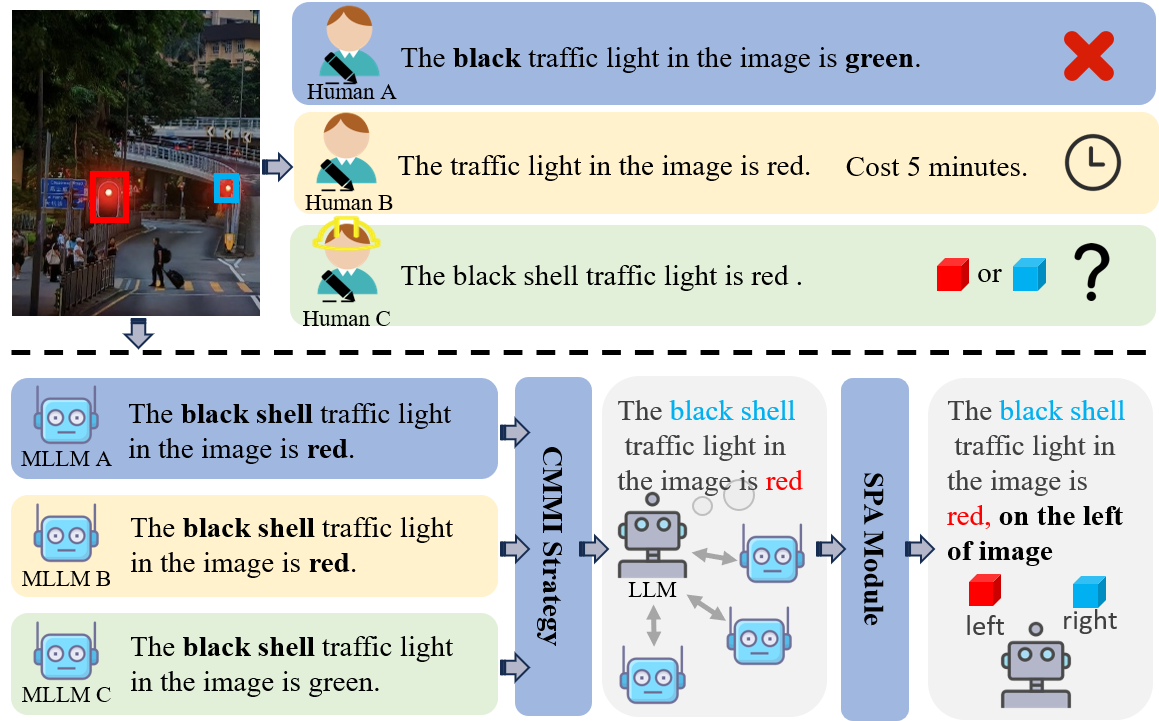}}
\caption{Comparison between manual annotation and ColLab. Manual annotation can be time-consuming and prone to errors, while our method effectively mitigates these issues.}
\label{pipeline}
\end{figure}

Recent developments in MLLMs~\cite{recent_mllms_chen2024internvl} have opened up new possibilities for automatic data construction~\cite{mllms_construct_fu2025trustgeogen}. 
However, generating descriptions directly using MLLMs still faces many challenges.
On the one hand, descriptions generated by a single MLLM may fail to capture the relevant features of an instance. 
On the other hand, utilizing multiple MLLMs can provide more diverse information, but manually filtering the outputs becomes an efficiency bottleneck. Moreover, for the REC task, the description needs to be unique. If multiple instances of the same category in an image share identical descriptions, it will lead to ambiguity.

In this work, we propose \textbf{ColLab}, a collaborative spatial progressive data engine designed for fully automated REC and REG dataset construction without any human annotation. As shown in Fig.~\ref{pipeline}, ColLab increases the capabilities of MLLMs and LLMs through a novel CMMI strategy. CMMI coordinates multiple MLLMs to generate descriptions, which are then refined by LLMs. 
We introduce the Spatial Progressive Augmentation (SPA) module to further enhance the fine-grained distinguishability of referring expressions, especially in complex scenes with multiple similar objects. SPA enriches the spatial cues within descriptions by injecting structured spatial information. Furthermore our framework has also been partially integrated into the dataset generation process of the MARS2 Challenge \cite{xu2025mars2} on Multimodal Reasoning\footnote{\url{https://github.com/mars2workshop}} and shape the Lens \cite{yao2025lens} benchmark quality.
In summary, the main contributions of this paper are as follows:
\begin{enumerate}
    \item We propose \textbf{ColLab}, a collaborative spatial progressive data engine that enables fully automated REC and REG data generation without human annotation. By harnessing the complementary strengths of MLLMs and LLMs, ColLab achieves substantial improvements in annotation efficiency and scalability.

    \item We introduce a \textbf{Collaborative Multimodal Model Interaction (CMMI)} strategy that  leverages MLLMs to collaboratively generate diverse and rich descriptions. This strategy enhances data diversity and semantic expressiveness, establishing a solid foundation for subsequent fine-grained spatial modeling.

    \item We design a \textbf{Spatial Progressive Augmentation (SPA)} module, which progressively strengthens spatial cues within the descriptions. This module improves fine-grained distinguishability among instances within the same category.
\end{enumerate}

\section{Related Work}

\subsection{Referring Expression Comprehension and Generation}
Recent REC and
REG are two complementary tasks that establish the connection between visual content and natural language.
REC focuses on localizing a target region in an image based on a natural language description, while REG aims to generate a description that uniquely identifies a specific region or object in the image~\cite{reg_tanaka2019generating, antol2015vqa}. 
Most existing REC and REG benchmarks are heavily reliant on manual annotations. 
For instance, Flickr30K Entities dataset~\cite{plummer2015flickr30k_Entities} extends the original Flickr30K dataset~\cite{flickr30k_young2014image} by aligning textual phrases with manually labeled image regions.
ReferItGame dataset~\cite{kazemzadeh2014referitgame} and RefCOCO/RefCOCO+/RefCOCOg datasets~\cite{RefCOCO+_yu2016modeling, RefCOCOg_mao2016generation} are built upon MSCOCO dataset~\cite{MSCOCO_lin2014microsoft}, where descriptions are collected via interactive games and the corresponding regions are annotated with bounding boxes.
Despite their widespread usage, the creation of these datasets is time-consuming and labor-intensive ~\cite{labor_intensive_chen2023advancing, labor_intensive_ma2024visual, labor_intensive_xu2024mc}, which hinders scalability.

\subsection{Large Language Models}
LLMs are deep learning models trained on large-scale textual data, capable of understanding and generating natural language.
In recent years, with the advancement of computing power, a series of representative LLMs have emerged, such as the GPT~\cite{gpt4_achiam2023gpt} series, LLaMA~\cite{llama_touvron2023llama} and DeepSeek-V3~\cite{deepseekv3_liu2024deepseek}.
These models have demonstrated strong general capabilities across a wide range of natural language processing tasks.

LLMs have shown remarkable advantages in understanding textual semantics and extracting key information, and are widely applied in structured information output tasks. An increasing number of studies are leveraging the text understanding capabilities of LLMs to transform text into specified types of outputs based on prompts~\cite{llm_text2json_chen2025llmer,yao2025lens,xu2025mars2}.

\subsection{Multimodel Large Language Models}
Recent MLLMs have emerged as a unified framework capable of understanding and generating content in multiple modalities such as visual~\cite{MLLMs_naveed2023comprehensive} and audio~\cite{mllms_audio_kuan2025can}. Unlike traditional single-modality models, MLLMs are jointly trained on large-scale multimodal data, enabling them to learn semantic correspondences between different data types and perform a wide range of multimodal tasks, including image captioning, visual question answering, and multimodal retrieval.
Recent advances have introduced several representative MLLMs, demonstrating their remarkable ability to generate instance-level descriptions with rich semantic details. GPT-4V~\cite{gpt4v_yang2023dawn} extends the capabilities of GPT-4~\cite{gpt4_achiam2023gpt} to visual input, allowing free-form interaction between text and images. MiniGPT-4~\cite{minigpt4_zhu2023minigpt} and BLIP-2~\cite{blip2_li2023blip} focus on efficient visual-language alignment by freezing a visual encoder and aligning it with a large language model using lightweight adapters.
LLaVA~\cite{llava_liu2023visual} further enhances visual-language interaction through instruction tuning and visual grounding, allowing the model to follow complex multimodal prompts. Qwen2.5-VL~\cite{bai2025qwen2} also demonstrates very powerful visual-language multimodal capabilities.

\begin{figure*}[t]
\centerline{\includegraphics[width=1\textwidth]{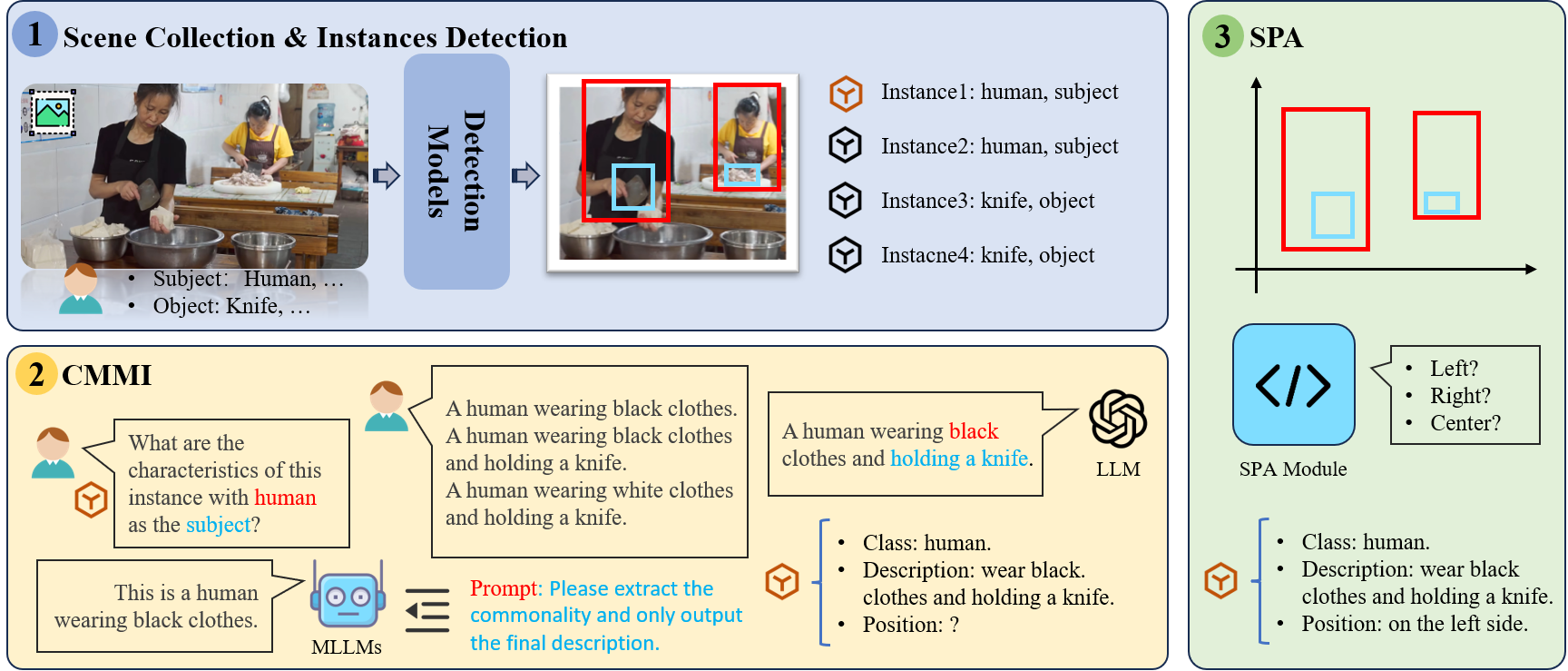}}
\caption{ColLab overview: \textbf{(1) Scene Collection and Instance Detection} collecting images and using detection models to generate instances. (2) \textbf{CMMI} strategy generates multiple descriptions for each instance and integrates them together. (3) \textbf{SPA} module adds spatial information to the generated descriptions.}
\label{fig2}
\end{figure*}

\section{Methodology}
The automated REC and REG data engine proposed in this paper consists of three key steps: 
scene collection and instance detection, collaborative multimodal model interaction, and spatial progressive augmentation. The overall framework is illustrated in Fig.~\ref{fig2}.

\subsection{Scene Collection and Instance Detection}\label{AA}
Firstly, we collected diverse scenes to ensure the richness of the dataset content as listed in Table~\ref{scenario_class}.
After collecting these scenes, images are fed into a detection model to automatically identify and generate multiple instances. Some instances generated by the detection model may contain recognition errors. In this case, we select only instances with a confidence score greater than 0.5 as input for the subsequent steps.

\begin{table}[t]
\caption{A collection of different types of scene.}
\begin{center}
\setlength\tabcolsep{10pt}
\begin{tabular}{cc}
\toprule
\textbf{Attribute} & \textbf{Scene} \\
\midrule
Outdoor & Street, Road, Playground, ...\\
Workplace & Airport, Train Station, Classroom, ... \\
Indoor & Kitchen, Bedroom, Living Room, Shower Room, ...\\
\bottomrule
\end{tabular}
\label{scenario_class}
\end{center}
\end{table}

\begin{table}[t]
\caption{Categorical division of subject and object types.}
\begin{center}
\setlength\tabcolsep{14.7pt}
\begin{tabular}{cc}
\toprule
\textbf{Attribute} & \textbf{Category} \\
\midrule
Subject & Human, Animal, Robot, Industrial Machine, ... \\
Object & Utensil, Container, Tool, Food, Clothing, ...  \\
\bottomrule
\end{tabular}
\label{subject_object}
\end{center}
\end{table}

Before the generation and extraction stages, each collected instance is associated with a predefined category label. 
We observe that instances from different categories exhibit distinct patterns in their feature descriptions.
To better capture these differences, we group categories into Subject and Object types shown in Table~\ref{subject_object}, enabling more structured and semantically-aware description generation.
For example, the category \textit{person} typically functions as a subject and frequently participates in actions involving other entities, as in \textit{a person holding a cup}, where person is the subject and cup is the object. These relational patterns reflect the inherent semantic roles of different categories.
Ignoring such distinctions can lead to inaccurate or semantically incomplete descriptions. 
However, in scenarios with limited interaction between objects, the Subject and Object types may not be able to fully demonstrate their effectiveness.

Finally, image regions corresponding to the generated categories are cropped based on the bounding boxes, and the previously defined Subject and Object types are appended as prompts. These serve as essential inputs for the subsequent processing stages.

\subsection{Collaborative Multimodal Model Interaction Strategy}
After obtaining the instance image and instance attributes mentioned in the previous steps, it will be used as a prompt and fed into multiple MLLMs.
Each MLLM generates a description for the instance based on the cropped region and a provided prompt, reflecting its unique understanding of the visual content.
However, due to the varying understanding capabilities between different MLLMs, the descriptions for the same instance may contain detailed errors.
To generate correct instance descriptions, LLMs are used to analyze and fuse the multiple candidate descriptions produced by different MLLMs.
This direct collaborative approach using LLMs can effectively extract the correct description details from different MLLMs and merge them into a final description.

\subsection{Spatial Progressive Augmentation Module}

\begin{figure}[t]
\centerline{\includegraphics[width=0.8\linewidth]{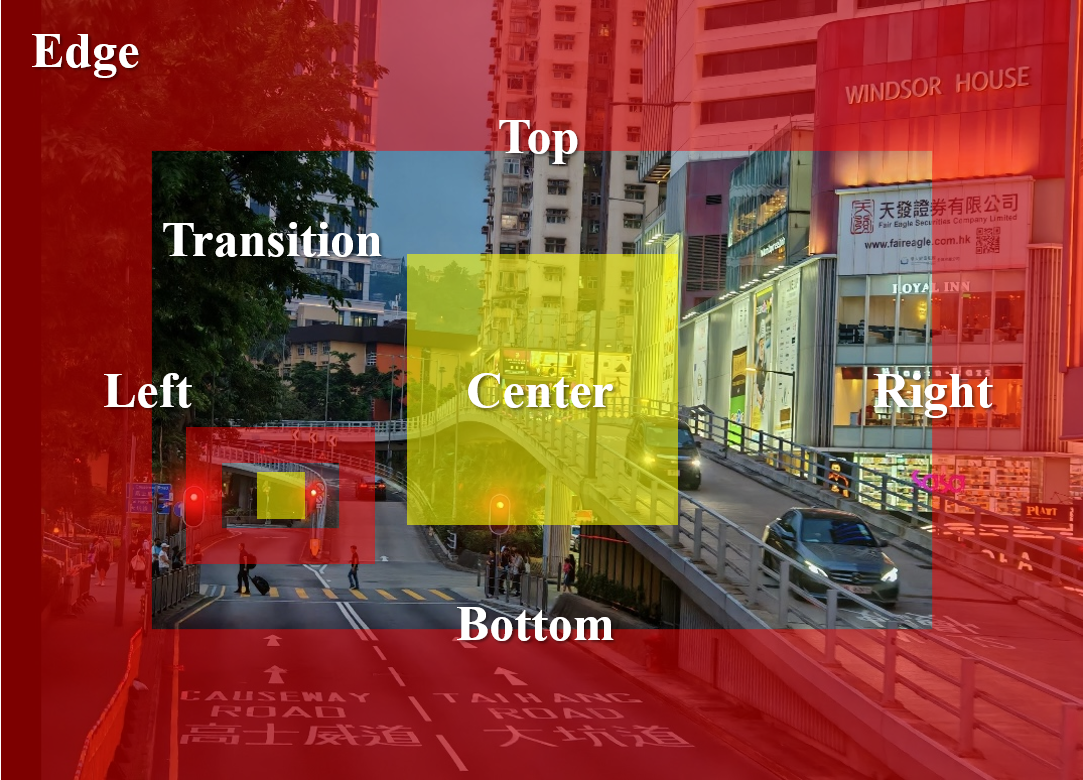}}
\caption{The division of regions within the same image, as well as the finer-grained subdivisions within each region.}
\label{spatial_augmentation}
\end{figure}

\IncMargin{1em}
\begin{algorithm}
\label{alg:spatial_augmentation}
    \SetAlgoLined
    \caption{Spatial Progressive Augmentation }
    \SetKwInOut{Input}{Input}
    \SetKwInOut{Output}{Output}

    \Input{Instance set $S$ containing \{$x_{min}$, $y_{min}$, $x_{max}$, $y_{max}$, $c$, $d$\}}
    \Output{Augmented instance descriptions}

    Initialize duplicated description set $K \leftarrow \emptyset$\

    \For{$I_i$ in $S$}{
        \If{$\text{GroupBy}(I_i[c],I_i[d])$ \textgreater \text{1}}{
            Add groups to $K$\
        }
    }

    \For{$k_i$ in $K$}{
        Initialize spatial region set $H \leftarrow \emptyset$\
        
        \For{$I_j$ in $k_i$}{
            Determine spatial region based on \{$x_{min}$, $y_{min}$, $x_{max}$, $y_{max}$\}\
            Assign instance to corresponding region $h_j$ in $H$\
        }

        \For{$h_j$ in $H$}{
            \If{number of instances in $h_j \geq 2$}{
                Further subdivide $h_j$ into finer spatial regions\
            }
        }

        Update instance descriptions with spatial terms based on their final region\
    }

    \KwRet Augmented instance descriptions
\end{algorithm}
\DecMargin{1em}

Unlike REG datasets, the instance description in the REC datasets format must be unique. So we propose an algorithm named Spatial Augmentation to distinguish instances with identical spatial description.
As illustrated in Fig.~\ref{spatial_augmentation}, the image is divided into three regions (center, transition and edge) and four directional areas (top, bottom, left and right).
According to Algorithm~\ref{alg:spatial_augmentation}, we first process instances within the same image that share identical descriptions. Then, spatial information is initially assigned based on the predefined regions. However, if multiple instances are assigned to the same region, we apply a recursive strategy: the region is subdivided into smaller sections using the same method outlined in Fig.~\ref{spatial_augmentation}, ensuring each instance is allocated to a distinct subregion. The spatial information will be reassigned accordingly.

\section{Experiment}
\subsection{Experimental Setup}
To evaluate the effectiveness of ColLab, we conduct an experiment using $N$ different MLLMs and a single LLM through API access. Detailed setting is provided in Step~\ref{E_C}.
We collect a total of 100 images to ensure semantic richness and scene complexity. These images cover various categories listed in Table~\ref{sence}. Moreover, we can select domain-specific detection models, MLLMs, and LLMs for diverse scenarios, significantly expanding our strategy applicability to broader contexts.

\begin{table}[t]
\caption{Statistics on the number of collected scenes.}
\begin{center}
\setlength\tabcolsep{14pt}
\begin{tabular}{cccc}
\toprule
\textbf{Scene} & \textbf{Amount}& \textbf{Scene}& \textbf{Amount} \\
\midrule
Street & 17 & Train Station & 4\\
Airport & 14 & Library & 4\\
Kitchen & 8 & Living Room & 14\\
Restaurant & 5 & Road & 1\\
Bedroom & 17 & Classroom & 13\\
Playground & 1 & Shower Room & 2\\
\bottomrule
\end{tabular}
\label{sence}
\end{center}
\end{table}

We employ object detection models \cite{yolo_jiang2022review,yao2024ctod,bai2025qwen2} to extract $K$ instances per image and retain only those with a confidence score above 0.5 to ensure detection quality.
As a result, $K=574$ instance bounding boxes are generated across all collected images.
Each cropped instance region is fed into a feature extraction network to obtain a two-dimensional feature embedding, as shown in Fig.~\ref{crop_distribution}. It can be observed that the \textit{person} category constitutes the majority of detected instances and their features are distributed within a relatively compact region in the embedding space.

\begin{figure}[t]
\centerline{\includegraphics[width=\linewidth]{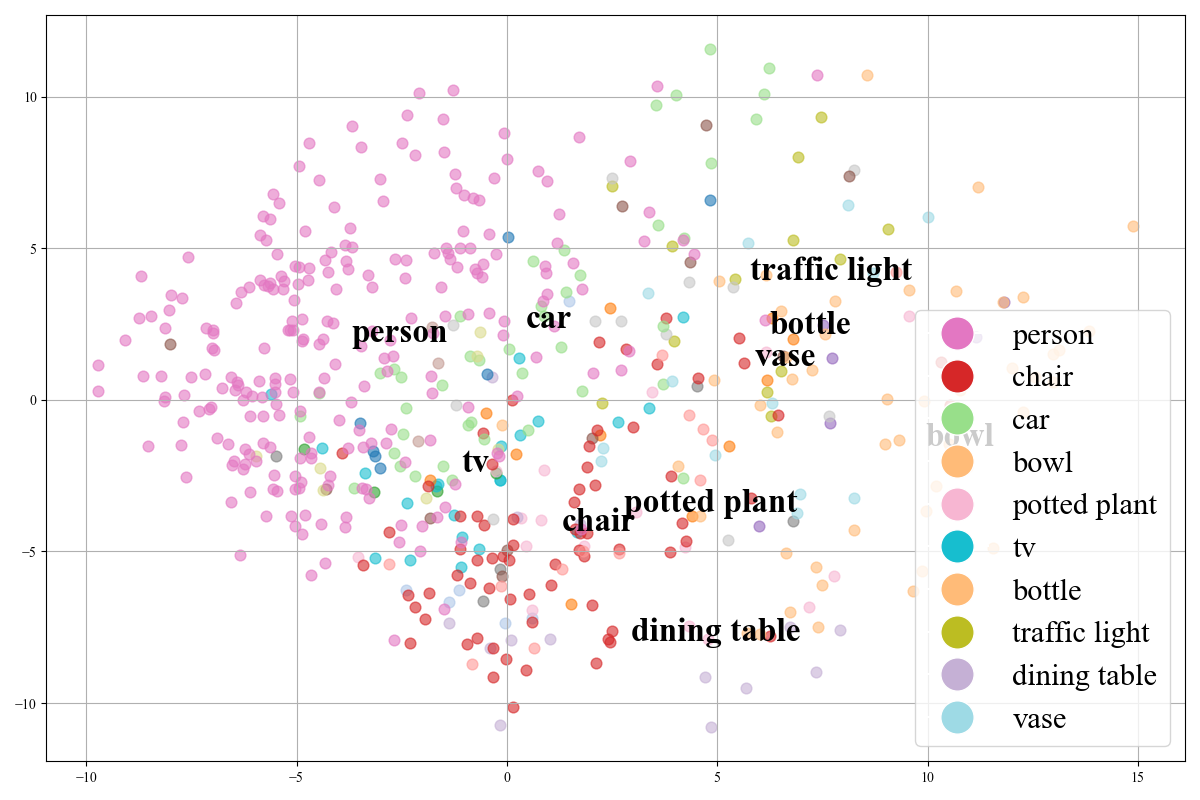}}
\caption{Visualization of 2D feature distribution for instance images cropped from various scenes, highlighting the top 10 categories by frequency.}
\label{crop_distribution}
\end{figure}

\begin{table}[t]
\caption{Data Statistics of Descriptions Generated by Different MLLMs Based on 
instance image.}
\begin{center}
\setlength\tabcolsep{13.5pt}
\begin{tabular}{cccc}
\toprule
\textbf{Model} & \textbf{Length (n)} & \textbf{Var (n)} & \textbf{Time (s)} \\
\midrule
Qwen2.5-VL-3B  & 16.32  & 12.88 & 3.47 \\
Qwen2.5-VL-7B  & 20.68  & 18.31 & 3.81 \\
Qwen2.5-VL-32B & 26.54 & 23.80 & 4.71 \\
\bottomrule
\end{tabular}
\label{MLLM_statistics}
\end{center}
\end{table}

\begin{figure*}[t]
\centerline{\includegraphics[width=\textwidth]{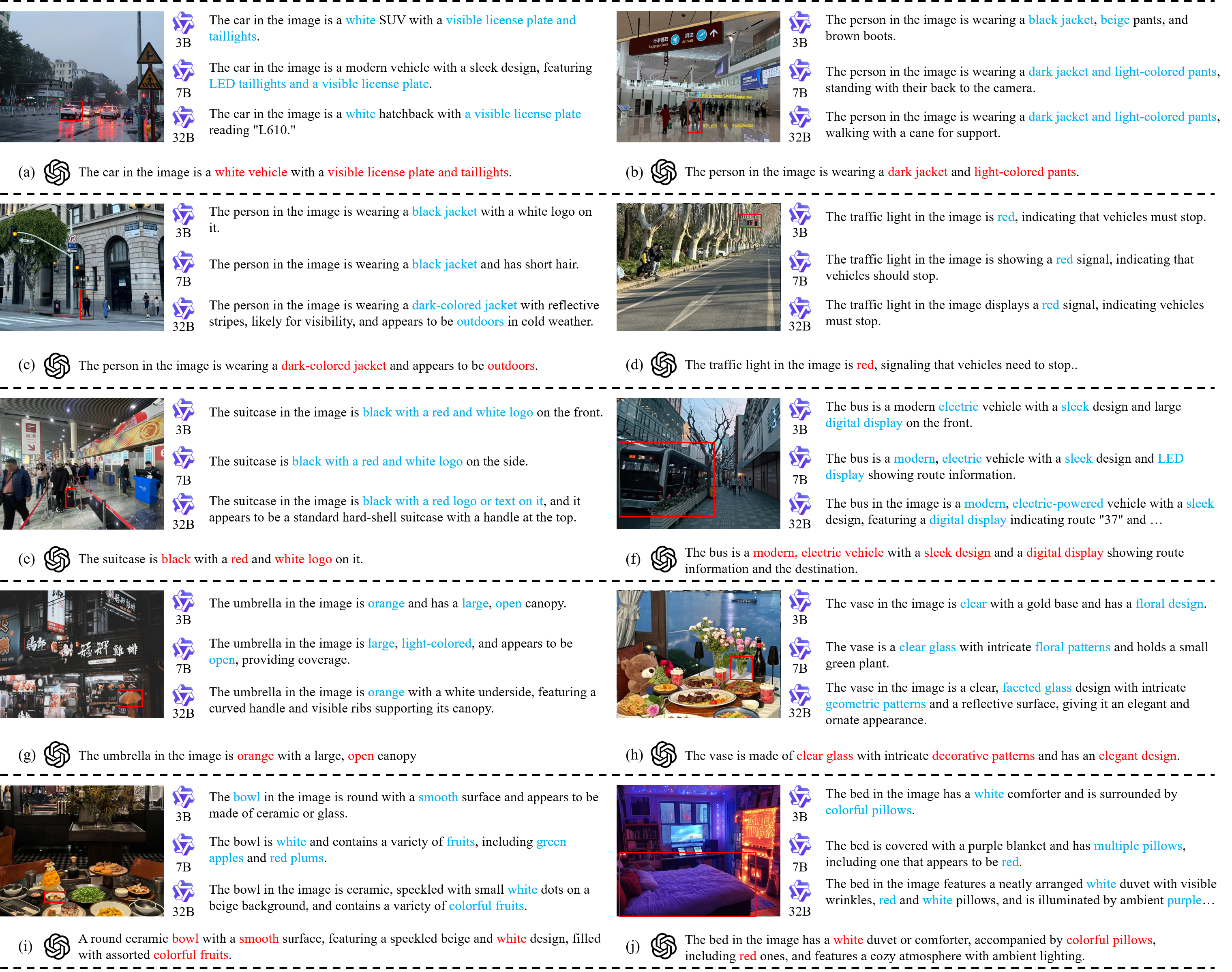}}
\caption{Some instance examples from different scenes, showing the descriptions generated by three different MLLMs and the common features extracted through the LLM.}
\label{example_aug}
\end{figure*}

\subsection{Performance of Collaborative Multimodal Model Interaction Strategy}
\subsubsection{\textbf{Generated Descriptions of MLLMs with Different Scales}}\label{E_C}
To evaluate the collaborative capability and complementarity of different MLLMs, we first analyze the descriptive characteristics of each individual model across different parameter scales. 
We set $N = 3$ and employ three different models: Qwen2.5-VL-3B, Qwen2.5-VL-7B, and Qwen2.5-VL-32B, generating a total of $K \times N$ instances. 
We select models with varying parameter scales to explore their differences in understanding image content. These MLLMs exhibit diverse capabilities in interpreting image instances and enriching the generated descriptions.
As illustrated in Fig.~\ref{example_aug}(d), the MLLMs not only extract color attributes of the traffic light but also capture its semantic implication, such as \textit{signaling that vehicles need to stop}. This highlights the capacity of MLLMs to generate context-aware and semantically rich instance-level descriptions.

To ensure the consistency and correctness of the MLLMs' outputs, we use a fixed prompt template:
\begin{equation}
 \mathrm{What~are~the~characteristics~of~\textit{C}~in~\textit{$I_c$}?}
\end{equation}
Where $C$ and $I_c$ denote the instance category and the instance image, respectively.
Following the generation process, $K\times N$ instance-level descriptions can be obtained. 
For a quantitative evaluation, we evaluate the model from three aspects.

\begin{itemize}[leftmargin=1.5em]
  \item[1)] \textbf{Average description length (Length):} Measuring the mean number of words per generated description.
  \item[2)] \textbf{Description length variance (Var):} Captureing variability across different instances.
  \item[3)] \textbf{Average description generation time (Time):} Evaluating computational efficiency of description generation.
\end{itemize}

The detailed statistics are presented in Table~\ref{MLLM_statistics}.
Based on the results in Table~\ref{MLLM_statistics}, we observe that the Qwen2.5-VL-32B model outperforms the other models in both average description length and variance of description length. Specifically, Qwen2.5-VL-32B generates the longest descriptions on average (26.54 words) and exhibits the highest variability in description length (23.80 words). This suggests that the large-scale model tends to produce more detailed and diverse descriptions compared to the smaller models, such as Qwen2.5-VL-3B and Qwen2.5-VL-7B, which generate shorter and more uniform descriptions. 
These results indicate that using diverse MLLMs can effectively mitigate  the issue of generating monotonous descriptions from a single MLLM.

\begin{figure}[t]
\centerline{\includegraphics[width=0.9\linewidth]{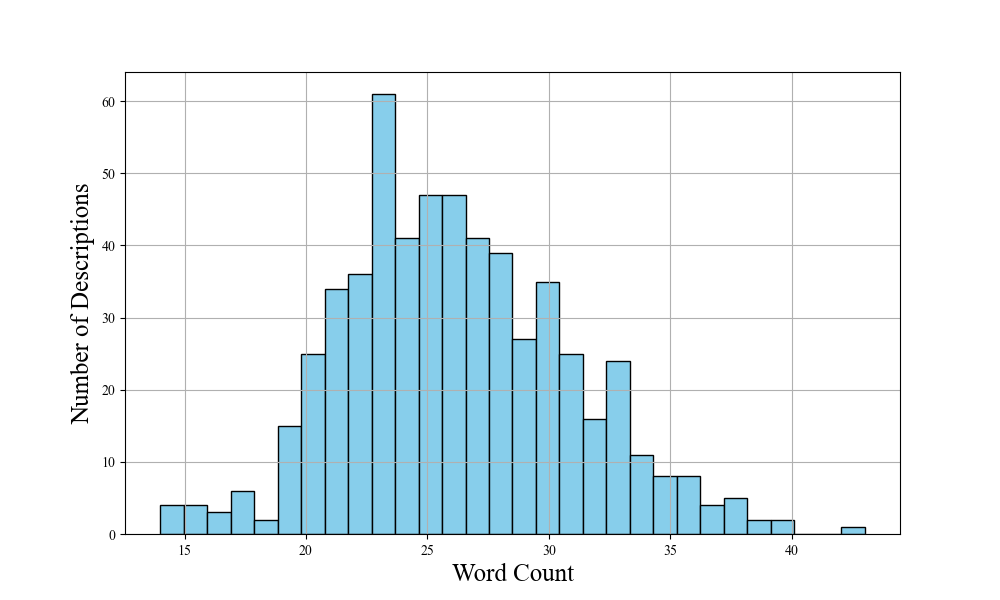}}
\caption{Distribution of word counts in descriptions generated by Qwen2.5-VL-32B.}
\label{qwen32b_static}
\end{figure}

To further analyze the descriptive capacity of Qwen2.5-VL-32B, we conduct a fine-grained statistical analysis by counting the number of generated descriptions for each word count.
The distribution is visualized in Fig.~\ref{qwen32b_static}. It can be seen that most descriptions are fall within the range of 20 and 30 words, with a peak around 24–26 words. This suggests that large-scale model tends to generate moderately detailed sentences, providing a good balance between richness and brevity.

These results confirm that large-scale MLLMs are capable of generating more diverse and informative descriptions. To further exploit the complementary strengths of different models, we next perform instance-level description fusion through a collaborative strategy.

\subsubsection{\textbf{LLMs-based Descriptions Merging}}
To realize collaborative interaction among multiple multimodal models, we employ Deepseek-V3 to perform description fusion and refinement. Specifically, for each instance region cropped from the raw image, we aggregate the descriptions generated by three different MLLMs (Qwen2.5-VL-3B, 7B, and 32B) and input them into Deepseek-V3 for processing.
A standardized prompt is designed to guide Deepseek-V3 in identifying the commonalities across the input descriptions and generating a coherent, instance-level expression. The output is formulated in a style consistent with the REC and REG datasets. The prompt is defined as follows:
\begin{equation}
    \mathrm{
    Extract~descriptions~of~\textit{C}~based~on~\textit{D}=\{d_1, \dots, d_i\}
    }
\end{equation}
where $C$ represents the category of the instance and $D$ denotes the set of descriptions generated by MLLMs.

Through this method, Deepseek-V3 effectively summarizes the key features shared across the input descriptions and generates a final instance-level description.
To intuitively illustrate the effectiveness of this extraction and generation process, we visualize several representative examples. As shown in Fig.~\ref{example_aug}, we present the original three descriptions for each instance, alongside the final description generated by Deepseek-V3. 
As shown in Fig.~\ref{example_aug}(a), two of the descriptions include the keyword \textit{white}, and all three descriptions mention \textit{plate} and \textit{taillights}.
The final expression generated by the Deepseek-V3 retains these common features. Notably, the 32B model even captures content from the license plate.
However, due to its low occurrence frequency, this detail is discarded in the final merged description.
It can be observed that the final descriptions are more relevant and consistent, demonstrating the ability of LLMs to perform effective description fusion.

In summary, the above observations demonstrate that the CMMI strategy effectively captures instance features and generates more informative descriptions.
If online latency is taken into consideration, a single-threaded linear processing approach can handle up to 5,082 items per day. Using multithreading would yield even better performance.

\begin{figure}[t]
\centerline{\includegraphics[width=\linewidth]{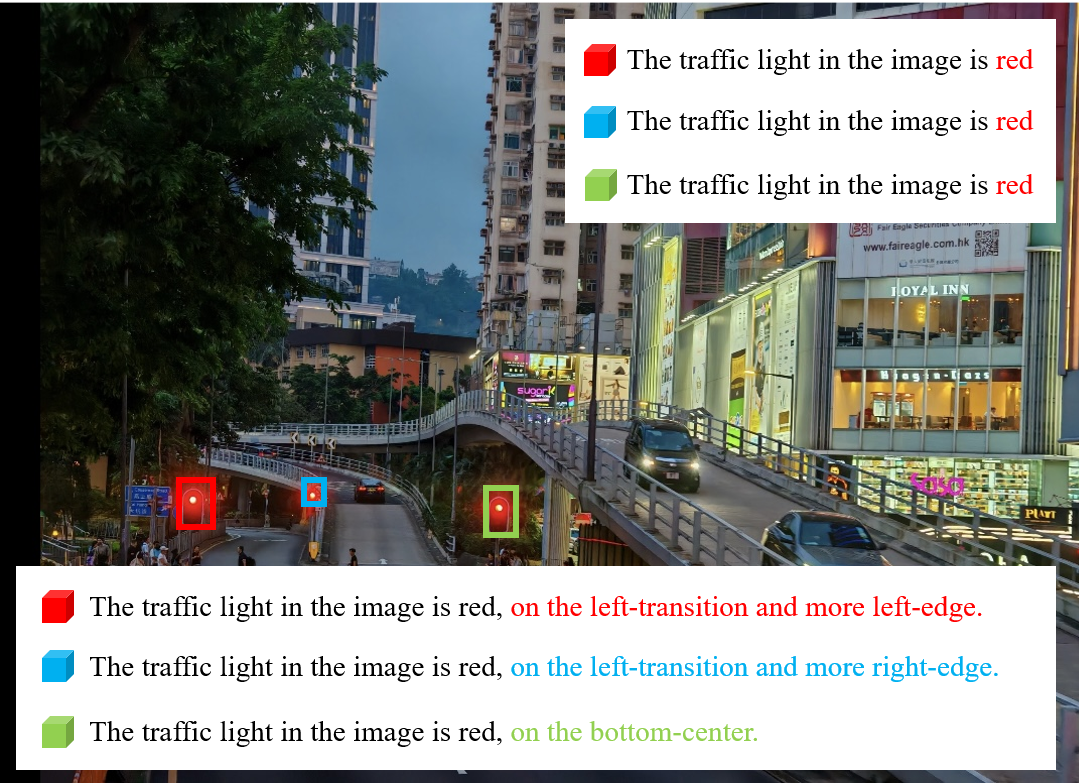}}
\caption{An example where three instances generated identical descriptions, along with the results after enhancing the descriptions using the SPA module.}
\label{ex_SPA}
\end{figure}

\subsection{Advantages Provided by Spatial Progressive Augmentation Module}
To improve the discriminability of descriptions for multiple instances of the same category within a single image, we evaluate the effectiveness of the proposed SPA module.
In Section~\ref{E_C}, identical descriptions may appear for images of the same category within a single image. As shown in Fig.~\ref{ex_SPA}, the three bounding boxes correspond to three instances of traffic lights. However, they all share an identical description: \textit{The traffic light in the image is red}. Clearly, such data is not suitable for use in REC tasks. 

To address this, we enhanced the spatial information of the three instances using the SPA module. 
As described in Section~\ref{E_C}, it is common for different instances of the same category in a single image to share identical descriptions. For example, as shown in Fig.~\ref{ex_SPA}, three traffic light instances are enclosed within bounding boxes but all receive the same expression: \textit{The traffic light in the image is red}. Such redundancy limits the utility of these descriptions for REC tasks, which require instance-level specificity. To address this issue, the SPA module introduces hierarchical spatial differentiation.
In the initial partitioning step, the green instance is assigned to the lower part of the \textit{bottom-center} region, while the red and blue instances are placed in the \textit{left-transition} region. Since the green instance is uniquely located, it is excluded from subsequent spatial refinement. The red and blue instances are grouped into a new region, where the spatial refinement process is repeated. As a result, the red instance is finally allocated to the \textit{left-edge} region, and the blue instance to the \textit{right-edge} region.

Following this process, the three instances receive spatially distinctive descriptions, enabling the generation of valid REC and REG training data. These results demonstrate that the SPA module effectively enhances spatial granularity and resolves semantic duplication in multi-instance settings.

\section{Conclusions}
In this work, we propose an automated framework named ColLab for constructing referring expression comprehension and referring expression generation datasets via a collaborative spatial progressive data engine. Our framework eliminates the need for manual annotation, enabling scalable dataset generation with minimal human intervention.
Moreover, we introduce a Collaborative Multimodal Model Interaction (CMMI) strategy that effectively generates high-quality instance descriptions directly from raw images, enhancing the richness of the dataset. To further address the challenge of identical generated descriptions, we design an Spatial Progressive Augmentation (SPA) module that significantly improves the diversity and specificity of generated descriptions. Extensive analysis shows that ColLab provides a reliable foundation for efficiently building referring expression comprehension and referring expression generation datasets with quality. In future, we will adopt larger-scale multimodal models and detection models adaptable to various scenarios to further enhance the quality and efficiency of the constructed data. Our framework was partially adopted in the data generation pipeline of the ICCV 2025 MARS2 Challenge and enrich the Lens benchmark with diverse, domain-oriented samples. This integration not only enhanced the dataset’s difficulty and representativeness but also validated the scalability of our approach through its influence on widely used multimodal reasoning benchmarks.

\printbibliography

\end{document}